\documentclass[letterpaper, 10pt, conference]{ieeeconf} 
\IEEEoverridecommandlockouts                          
\makeatletter
\let\NAT@parse\undefined
\makeatother

\usepackage{times}
\usepackage[numbers]{natbib}
\usepackage{multicol}
\usepackage{hyperref}
\usepackage[english]{babel}								
\usepackage[utf8]{inputenc}
\usepackage[T1]{fontenc}								
\usepackage{graphicx,subfig,color,epsfig} 				
\usepackage{amsmath,amssymb} 
\usepackage{todonotes}									
\usepackage{lineno}										
\usepackage{paralist}									
\usepackage{multirow}									
\usepackage{float}
\usepackage{epstopdf}

\newcommand{\ruben}[1] {\textcolor{black}{#1}}

\title{ Learning-based Image Enhancement for Visual Odometry in Challenging HDR Environments }

\author{Ruben Gomez-Ojeda$^{1}$, Zichao Zhang$^{2}$, Javier Gonzalez-Jimenez$^{1}$, Davide Scaramuzza$^{2}$
\thanks{ $^{1}$R. Gomez-Ojeda and J. Gonzalez-Jimenez are with the Machine Perception and Intelligent Robotics (MAPIR) Group, University of Malaga, Spain. (email: rubengooj@gmail.com, javiergonzalez@uma.es). 
	\url{http://mapir.isa.uma.es/}.
}
\thanks{ $^{2}$Z. Zhang and D. Scaramuzza are with the Robotics and Perception Group, Dep. of Informatics, University of Zurich, and Dep. of Neuroinformatics, University of Zurich and ETH Zurich, Switzerland. 
	(email: {zzhang,sdavide}@ifi.uzh.ch)
	\url{http://rpg.ifi.uzh.ch.}
}
\thanks{This work has been supported by the Spanish Government (project DPI2014-55826-R and grant BES-2015-071606).}
}

\begin{document}

\maketitle
\thispagestyle{empty}
\pagestyle{empty}

\begin{abstract}
One of the main open challenges in visual odometry (VO) is the robustness to difficult illumination conditions or high dynamic range (HDR) environments.
The main difficulties in these situations come from both the limitations of the sensors and the inability to perform a successful tracking of interest points because of the bold assumptions in VO, such as brightness constancy.
We address this problem from a deep learning perspective, for which we first fine-tune a deep neural network 
with the purpose of obtaining enhanced representations of the sequences for VO.
Then, we demonstrate how the insertion of long short term memory allows us to obtain temporally consistent sequences, as the estimation depends on previous states.
%
\ruben{However, the use of very deep networks enlarges the computational burden of the VO framework; therefore, we also propose a convolutional neural network of reduced size capable of performing faster.}
%
Finally, we validate the enhanced representations by evaluating the sequences produced by the two architectures in several state-of-art VO algorithms, such as ORB-SLAM and DSO.
\end{abstract}

\newcommand {\indfiga}{-6mm}							
\newcommand {\indfigb}{-6mm}							
\newcommand {\indfigc}{-6mm}							
\newcommand {\indfigd}{-6mm}							

\newcommand {\indtaba}{-3mm}							
\newcommand {\indtabb}{-2mm}							
\newcommand {\indtabc}{-3mm}							

\newcommand {\indfig}{-6mm}							
\newcommand {\indtab}{-2mm}							
\newcommand {\indtabbig}{-6mm}						
\newcommand {\indsubfig}{0mm}

\newcommand{\wintro}      {0.45\textwidth}				
\newcommand{\wteaser}     {0.45\textwidth}				

\newcommand{\wdnn}        {0.65\textwidth}				
\newcommand{\wdnnour}     {0.375\textwidth}				
\newcommand{\wtraintsu}   {0.18\textwidth}				
\newcommand{\wtrainurb}   {0.16\textwidth}				
\newcommand{\wresults}    {0.11\textwidth}				

\newcommand{\symGrad}{{g}}								
\newcommand{\symGradMean}{\mu_{\symGrad}}				
\newcommand{\symGradStd}{\sigma_{\symGrad}}			


\newcommand{\fig}[1]{Figure \ref{#1}}					
\newcommand{\figs}[2]{Figures \ref{#1} and \ref{#2}}	
\newcommand{\tab}[1]{Table \ref{#1}}					
\newcommand{\secref}[1]{Section {\ref{#1}}}				

\newcommand{\bs}[1]{\boldsymbol{#1}}					
\newcommand{\ssl}[1]{\tensor[^{#1}]}					
\newcommand{\MatrixS}[1]{\bs{#1}}					    
\newcommand{\MatrixL}[1]{\textbf{#1}}					
\newcommand{\brackets}[1]{\begin{bmatrix}#1\end{bmatrix}}		

\newcommand{\argmin}[1]{\underset{#1}{\operatorname{argmin}}}		
\newcommand{\argmax}[1]{\underset{#1}{\operatorname{argmax}}}		
\newcommand{\der}[2]{\frac{\partial #1}{\partial #2}}			
\newcommand{\derin}[3]{\left.\der{#1}{#2}\right|_{#3}} 			
\newcommand{\prob}[2]{p\left( #1 | #2 \right)}				
\newcommand{\norm}[1]{\left\lVert#1\right\rVert} 			
\newcommand{\fnorm}[1]{\left\lVert#1\right\rVert_{\mathfrak{F}}}	
\newcommand{\skewmat}[1]{ \left[#1\right]_\times }			

\newcommand{\IdMat}{\MatrixL{I}}					
\newcommand{\canonicalvec}[1]{\textbf{e}_{#1}}				
\newcommand{\TRANSPOSE}{^\top}						
\newcommand{\symcov}{\MatrixS{\Sigma}}					
\newcommand{\symRe}{\mathbb{R}}						
\newcommand{\symSSpace}{S}						
\newcommand{\symPSpace}{\mathbb{P}}					
\newcommand{\symRotSpace}{SO(3)}					
\newcommand{\symRotLie}{\mathfrak{so}(3)}				
\newcommand{\symEucSpace}{SE(3)}					
\newcommand{\symEucLie}{\mathfrak{se}(3)}				
\newcommand{\symrot}{\textbf{R}}					
\newcommand{\symtrans}{\textbf{t}}					

\newcommand{\idL}{L}				
\newcommand{\idR}{R}				
\newcommand{\idF}{k}				
\newcommand{\idFn}{k+1}				

\newcommand{\lIm}{\textbf{l}}			
\newcommand{\normL}{\eta_l}			
\newcommand{\lImFirst}{\lIm_{\idL,\idF}}	
\newcommand{\lImSecond}{\lIm_{\idR,\idF}}	
\newcommand{\lImThird}{\lIm_{\idL,\idFn}}	
\newcommand{\lImFourth}{\lIm_{\idR,\idFn}}	

\newcommand{\match}{\textbf{m}}			

\newcommand{\cam}{C}				
\newcommand{\calib}{\MatrixL{K}}		
\newcommand{\LO}{LO}				
\newcommand{\spoint}{\textbf{p}}		
\newcommand{\epoint}{\textbf{q}}		
\newcommand{\Spoint}{\textbf{P}}		
\newcommand{\Epoint}{\textbf{Q}}		
\newcommand{\spointx}{p_x}		
\newcommand{\spointy}{p_y}		

\newcommand{\separam}{\bs{\xi}}
\newcommand{\separaminc}{\bs{\varepsilon}}
\newcommand{\separamopt}{\separam^*}
\newcommand{\reltrans}{\MatrixL{T}(\separam)}
\newcommand{\reltransopt}{\MatrixL{T}(\separamopt)}

\newcommand{\symover}{\gamma}
\newcommand{\errfun}{\MatrixL{E}}
\newcommand{\weifun}{\MatrixL{W}}
\newcommand{\jacfun}{\MatrixL{J}}

\newcommand{\cross}{\times}	
\newcommand{\lx}{(p_y-q_y)}
\newcommand{\ly}{(q_x-p_x)}
\newcommand{\lbeta}{(p_x^2+p_y^2+q_x^2+q_y^2Id)-2(p_xq_x+p_yq_y}

\ruben{
\vspace{-2ex}
\section*{Supplementary Materials}
A video demonstrating the proposed method is available at \url{https://youtu.be/NKx_zi975Fs}.}

\section{Introduction}
\label{sec_introduction}
In recent years, Visual Odometry (VO) has reached a high maturity and there are many potential applications, such as unmanned aerial vehicles (UAVs) and augmented/virtual reality (AR/VR).
Despite the impressive results achieved in controlled lab environments, the robustness of VO in real-world scenarios is still an unsolved problem.
While there are different challenges for robust VO (e.g., weak texture  \cite{Eade2009}\cite{gomez2016pl}), in this work we are particularly interested in improving the robustness in HDR environments.
%
%
The difficulties in HDR environments come not only from the limitations of the sensors
(conventional cameras often take over/under-exposed images in such scenes), but also from the bold assumptions of VO algorithms, such as brightness constancy.
%
%
%
%
To overcome these difficulties, two recent research lines have emerged respectively: Active VO and Photometric VO.
The former  tries to provide the robustness by controlling the camera parameters (gain or exposure time) \cite{shim2014auto}\cite{zhang16active},
%
while the latter explicitly models the brightness change using the photometric model of the camera \cite{li2016hdrfusion} \cite{engel2016direct}.
These approaches are demonstrated to improve robustness in HDR environments. However, they require a detailed knowledge of the specific sensor and a heuristic setting of several parameters, which cannot be easily generalized to different setups.
%
%
%


%
In contrast to previous methods, we address this problem from a \textit{Deep Learning} perspective, taking advantage of the generalization properties to achieve robust performance in varied conditions.
Specifically, in this work, we propose two different Deep Neural Networks (DNNs) 
that enhance monocular images to more informative representations for VO.
Given a sequence of images, our networks are able to produce an enhanced sequence that is invariant to illumination conditions or robust to HDR environments and, at the same time, contains more gradient information for better tracking in VO.
%
%
For that, we add the following contributions to the state of the art:
\begin{enumerate}[$\circ$]
\item 
We propose two different deep networks: a very deep model consisting of both CNNs and LSTM, 
\ruben{and another one of small size designed for less demanding applications.}
Both networks transform a sequence of RGB images into more informative ones, while also being robust to changes in illumination, exposure time, gamma correction, etc.
%
%
\item 
We propose a multi-step training strategy that employs the down-sampled images from synthetic datasets, which are augmented with a set of transformations to simulate different illumination conditions and camera parameters.
As a consequence, our DNNs are capable of generalizing the trained behavior to full resolution real sequences in HDR scenes or under difficult illumination conditions.
\item Finally, we show how the addition of Long Short Term Memory (LSTM) layers helps to produce more stable and less noisy results in HDR sequences by incorporating the temporal information from previous frames. However, these layers increase the computational burden, hence complicating their insertion into a real-time VO pipeline.
\end{enumerate}
We validate the claimed features by comparing the performance of two state-of-art algorithms in monocular VO, 
namely ORB-SLAM \cite{mur2015orb} and DSO \cite{engel2016direct}, with the original input and the enhanced sequences, showing the benefits of our proposals in challenging environments.

\section{Related Work}
\label{sec_related}
%
To overcome the difficulties in HDR environments, works have been done to improve the image acquisition process as well as to design robust algorithms for VO.

\subsection{Camera Parameter Configuration}
The main goal of this line of research is to obtain the best camera settings (i.e., exposure, or gain) for image acquisition.
Traditional approaches are based on heuristic image statistics, typically the mean intensity (brightness) and the intensity histogram of the image.
For example, a method for autonomously configuring the camera parameters was presented in \cite{neves2009autonomous}, where the authors proposed to setup the exposure, gain, brightness, and white-balance by processing the histogram of the image intensity.
Other approaches exploited more theoretically grounded metrics.
\cite{lu2010camera}, employed the Shannon entropy to optimize the camera parameters in order to obtain more informative images. They experimentally proved a relation between the image entropy and the camera parameters, then selected the setup that produced the maximum entropy.

Closely related to our work, some researchers tried to optimize the camera settings for visual odometry.
\cite{shim2014auto} defined an information metric, based on the gradient magnitude of the image, to measure the amount of information in it, and then selected the exposure time that maximized the metric.
Recently, \cite{zhang16active} proposed a robust gradient metric and adjusted the camera setting according to the metric. They designed their exposure control scheme based on the photometric model of the camera and demonstrated improved performance with a state-of-art VO algorithm \cite{forster2014svo}.

%
%

\subsection{Robust Vision Algorithms}
To make VO algorithms robust to difficult light conditions, some researchers proposed to use invariant representations, while others tried to explicitly model the brightness change.
For feature-based methods, binary descriptors are efficient and robust to brightness changes. \cite{mur2015orb} used ORB features \cite{rublee2011orb} in a SLAM pipeline and achieved robust and efficient performance. Other binary descriptors \cite{Leutenegger11iccv}\cite{Calonder12pami} are also often used in VO algorithms.
For direct methods, \cite{alismail2016direct} incorporated binary descriptors into the image alignment process for direct VO, and the resulting system performed robustly in low light.

To model the brightness change, the most common technique is to use an affine transformation and estimate the affine parameters in the pipeline. \cite{jin2001real} proposed an adaptive algorithm for feature tracking, where they employed an affine transformation that modeled the illumination changes. More recently, a photometric model, such as the one proposed by \cite{debevec2008recovering}, is used to account for the brightness change due to the exposure time variation.
A method to deal with brightness changes caused by auto-exposure was published in \cite{li2016hdrfusion}, reporting a tracking and dense mapping system based on a normalized measurement of the radiance of the image (which is invariant to exposure changes). 
Their method not only reduced the drift of the camera trajectory estimation, but also produced less noisy maps.
\cite{engel2016direct} proposed a direct approach to VO with a joint optimization of both the model parameters, the camera motion, and the scene structure. They used the photometric model of the camera as well as the affine brightness transfer function to account for the brightness change.
%
%
In \cite{zhang16active}, the authors also adapted a direct VO algorithm \cite{forster2014svo} with both methods and presented an experimental comparison of using the affine compensation and the photometric model of the camera.

To the best of our knowledge, there is few work on using learning-based methods to tackle the difficulties in HDR environments.
In the rest of the paper, we will describe how to design networks for this task, the training strategy and the experimental results.
%
%
%
%

%
%


%
\begin{figure*}[!htb]
	\centering
	
	\subfloat[DNN model used in fine-tuning.]{	
		\includegraphics[width=\wdnn]{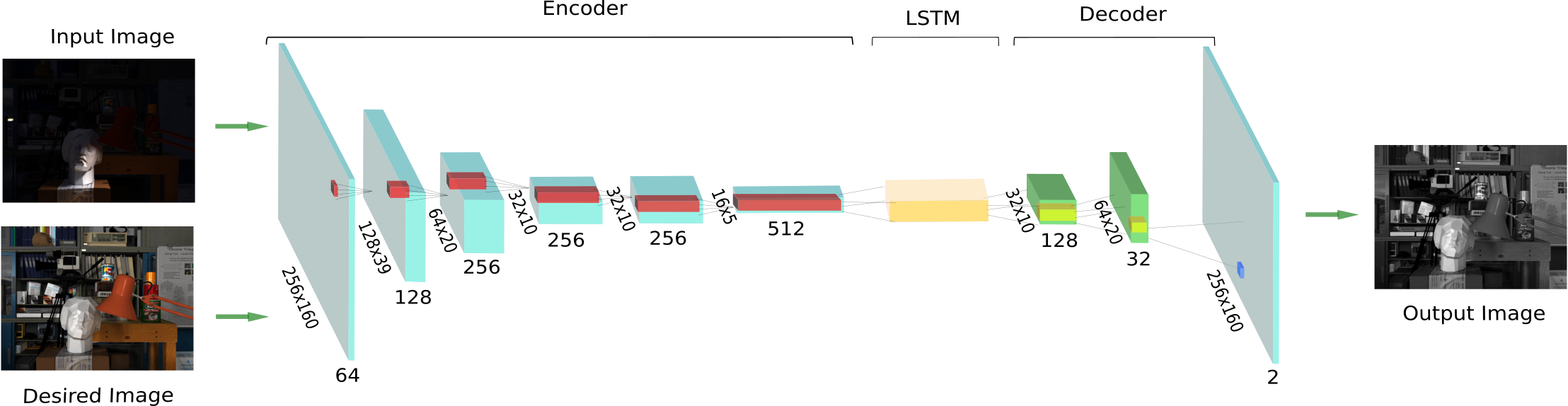}
		\label{finetune}
	}	
	\quad
	\subfloat[Small-CNN trained from scratch.]{	
		\includegraphics[width=\wdnnour]{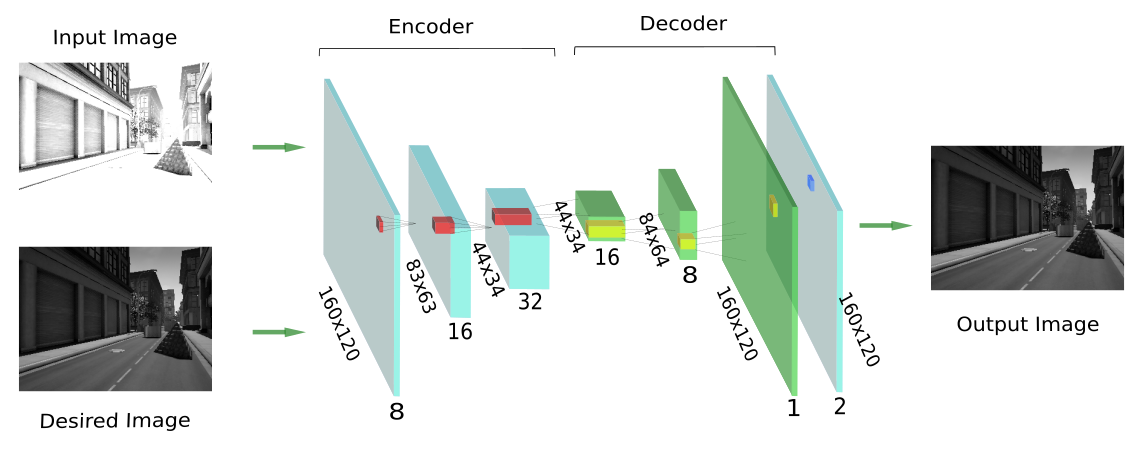}
		\label{proposal}
	}
	\caption{
		\textcolor{black}{
		Scheme of the architectures employed in this work. Both DNNs are formed by an \textit{encoder} convolutional network, and a \textit{decoder} that forms the enhanced output images. 
		In the case of the fine-tuned network, we introduce a LSTM network to produce temporally consistent sequences.
		%
		%
		These figures have been adapted from \cite{mancini2016fast, mancini16adomain}.
	}
	}
	\vspace{\indfiga}
	\label{fig_network}
\end{figure*}
\section{Network Overview}
\label{sec_overview}
%
%
In this work, we need to perform a pixel-wise transformation from monocular RGB images in a way that the outputs are still realistic images, on which we will further run VO algorithms.
For pixel-wise transformation, the most used approach is DNNs structured in the so-called \textit{encoder-decoder} form. 
These type of architectures have been successfully employed in many different tasks, such as optical flow estimation \cite{dosovitskiy2015flownet}, image segmentation \cite{kendall2015bayesian}, depth estimation \cite{mancini2016fast}, or even to solve the image-to-image translation problem \cite{isola2016image}.
The proposed architectures (see \fig{fig_network}), implemented in the Caffe library \cite{jia2014caffe}, consist of an encoder, LSTM layers and a decoder, as described in the following.

\subsection{Encoder}
The encoder network consists of a set of purely convolutional layers that transform the input image, into a more reduced representation of feature vectors, suitable for a specific classification task.
Due to the complexity of training from scratch \cite{tajbakhsh2016convolutional}, a standard approach is to initialize the model with the weights of a pre-trained model, known as \textit{fine-tuning}.
This has several advantages, as models trained with massive amount of natural images such as VGGNet \cite{simonyan2014very}, a seminal network for image classification, usually provide a good performance and stability during the training.
Moreover, as initial layers closer to the input image provide low-level information and final layers are more task-specific, it is also typical to employ the first layers of a well-trained CNN for different purposes, i.e. place recognition \cite{gomez2015training}.
This was also the approach in \cite{mancini16adomain}, where authors employed the first 8 layers of VGGNet to initialize their network, keeping their weights fixed during training, while the remaining layers were trained from scratch with random initialization.
Therefore, in this work, we first fine-tuned the very deep model in \cite{mancini16adomain}, depicted in \fig{finetune}.

\ruben{
However, since our goal is to estimate the VO with the processed sequences, a very deep network, such as the fine-tuned model,
is less suitable for usual robotic applications, where the computational power must be saved for the rest of modules.
}
%
Moreover, depth estimation requires a high level of semantic abstraction as it needs some spatial reasoning about the position of the objects in the scene.
In contrast, VO algorithms are usually based on tracking regions of interest in the images, which largely relies on the gradient, i.e., the first derivatives of the images, information that it is usually present in the shallow layers of CNNs.
%
\ruben{
Therefore, we also propose a smaller and less deep CNN to obtain faster performance, whose 
}
encoder is formed by three layers (dimensions are in \fig{proposal}), each one of them formed by a convolution with a $5\times 5$ kernel, followed by a batch-normalization layer \cite{ioffe2015batch} and a pooling layer.
%
%
%

\subsection{Long Short Term Memory (LSTM)}
%

While it is feasible to use a feedforward neural network to increase the information in images for VO, 
\ruben{the input sequence may contain non-ignorable brightness variation.
More importantly, the brightness constancy is not enforced in a feedforward network,
hence the output sequence is expected to break the brightness constancy assumption for many VO algorithms.
}
%
To overcome this, we can exploit the sequential information to produce more stable and temporally consistent images, i.e. reducing the impact of possible illumination change to ease the tracking of interest points.
Therefore, we exploit the Recurrent Neural Networks (RNNs), more specifically, the LSTM networks first introduced in \cite{hochreiter1997long}.
%
In these networks, unlike in standard CNNs where the output is only a non-linear function $f$ of the current state $\textbf{y}_t  = f(\textbf{x}_t)$, the output is also dependent on the previous output:
\begin{equation}
\textbf{y}_t  = f(\textbf{x}_t,\textbf{y}_{t-1})
\end{equation}
as the layers are capable of memorizing the previous states.
%
We introduce two LSTM layers in the fine-tuned network between the encoder and the decoder part, in order to produce more stable results for a better odometry estimation.

\subsection{Decoder}
%
Finally, the decoder network is formed by three deconvolutional layers, each of them formed by an upsampling, a convolution and a batch-normalization layer, as depicted in \fig{fig_network}.
The deconvolutional layers increase the size of the intermediate states and reduce the length of the descriptors.
%

\ruben{
Typically, decoder networks produce an output image of a proportional size of the input one containing the predicted values, which is in general blurry and noisy thus not very convenient to be used in a VO pipeline. 
To overcome this issue, we introduce an extra step which merges the raw output of the decoder with the input image producing 
a more realistic image.
For that, we concatenate both the input image in grayscale and the decoder output into a 2-channel image then applying a 
final convolutional filter with a $1\times 1$ kernel and one channel.
%
%
}

%

%
\section{Training the DNN}
\label{sec_training}
Our goal is to produce an enhanced image stream to increase the robustness/accuracy of visual odometry algorithms under challenging situations.
Unfortunately, there is no ground-truth available for generating the optimal sequences, nor direct measurement that indicates the goodness of an image for VO.
To overcome this difficulties, we observe that the majority of the state-of-art VO algorithms, both \textit{direct} and \textit{feature-based} approaches, actually exploit the gradient information in the image.
Therefore, we aim to train our network to produce images containing more gradient information.
In this section, we first introduce the dataset used for training then our training strategy.
%
%

\begin{figure}[!htb]
	\centering
	\subfloat[Reference]{	
		\includegraphics[width=\wtrainurb]{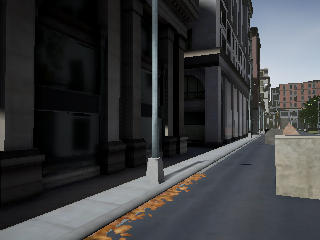}
		\label{urban_ref}
	}	
	\quad
	\subfloat[Dark conditions]{	
		\includegraphics[width=\wtrainurb]{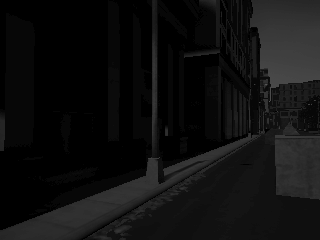}
		\label{urban_dark}
	}	
	\quad
	\subfloat[Daylight]{	
		\includegraphics[width=\wtrainurb]{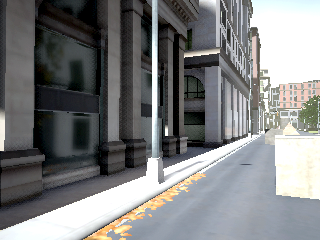}
		\label{urban_color}
	}
	\quad
	\subfloat[Over-exposition]{	
		\includegraphics[width=\wtrainurb]{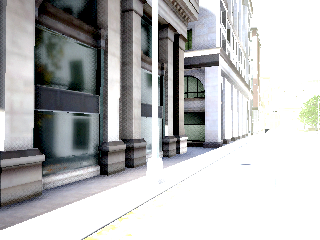}
		\label{urban_bright}
	}
	\caption{
		Some training samples from the Urban dataset proposed in \cite{mancini16adomain}, for which we have simulated artificial illumination and exposure conditions by post-processing the dataset with different contrast and gamma levels.
	}
	\vspace{\indfigb}
	\label{fig_urban_training}
\end{figure}

\subsection{Datasets}
To train the network, we need images taken at the same pose but with different illuminations, which are unfortunately rarely available in real-world VO datasets.
Therefore we employed synthetic datasets that contain changes in the illumination of the scenes.
In particular, we used the well-known New University of Tsukuba dataset  \cite{peris2012towards} and the Urban Virtual dataset generated by \cite{mancini16adomain}, consisting of several sequences from an artificial urban scenario with non-trivial 6-DoF motion and different illumination conditions.
%
\ruben{
In order to increase the amount of data, we simulated 12 different camera and illumination conditions (see \fig{fig_urban_training})
by using several combinations of Gamma and Contrast values.
Notice that this data augmentation must contain an equally distributed amount of conditions, otherwise the 
output of the network might be biased to the predominant case. 
%
%
%
}
%
To select the best image $\textbf{y}^*$ (with the most gradient information), we use the following gradient information metric:
\begin{equation}
\label{gradient_info}
\symGrad(\textbf{y}) =  \sum\limits_{ \textit{u}_i } \norm{\nabla \textbf{y} (\textit{u}_i)}^2
%
\end{equation}
which is the sum of the gradient magnitude 
over all the pixels $\textit{u}_i$ in the image $\textbf{y}$.
%
For training the CNN, we used 
RGB images of $256\times160$ pixels in the case of fine-tuning the model in \cite{mancini16adomain} and grayscale images of $160\times120$ pixels for the reduced network.
%
\ruben{
We trained the LSTM network with full-resolution images ($752\times480$) as, unlike convolutional layers, 
once trained they cannot be applied to inputs of different size.
}

\subsection{Training the CNN}
We first train without LSTM, with the aim of obtaining a good CNN (\textit{encoder}-\textit{decoder}) capable of estimating the enhanced images from individual (not sequential) inputs. This part of training consists of two stages:
%
\subsubsection{Pre-training the Network}
In order to obtain a good and stable initialization, we first train the CNN with pairs of images at the same pose, consisting of the reference image $y^*$ and an image with different appearance.
On our first attempts, we tried to optimize directly the bounded increments of the gradient information \eqref{gradient_info}. The results are very noisy, due to the high complexity of the pixel-wise prediction problem.
Instead, we opted to train the CNN by imposing the output to be similar to the reference image, in a pixel-per-pixel manner.
For that, we employed the logarithmic RMSE, which is defined for a given reference ${y}^*$ and an output ${y}$ image as:
\begin{equation}
\label{eq_logloss}
\mathcal{L}(\textbf{y},\textbf{y}^*) = \sqrt{ \frac{1}{N} \sum\limits_{i}  \norm{ \log\:{y}_i - \log\:{y}^*_i }^2 }   \;,
\end{equation}
where $i$ is the pixel index in the images.
%
\ruben{
Although we tried different strategies for this purpose, such as the denoising autoencoder \cite{vincent2008extracting},
we found this loss function much more suitable for VO applications, as it produced a smoother result than the Euclidean RMSE, 
specially for bigger errors, hence easing the convergence process.
}
%
This first part of the training was performed with the Adam solver \cite{kingma2014adam}, with a learning rate $l=0.0001$ for 20 epochs of the training data, and a dataset formed by 80k pairs and
requiring about 12 hours on a NVIDIA GeForce GTX Titan.

\subsubsection{Imposing Invariance}
Once a good performance with the previous training was achieved, we trained the CNN to obtain invariance to different appearances.
The motivation is that, for images with different appearances (i.e. brightness) taken at the same pose, the CNN should be able produce the same enhanced image.
For that, we selected triplets of images from the Urban dataset, by taking the reference image $\textbf{y}^*$, and another two images $\textbf{y}_1$ and $\textbf{y}_2$ from the same place with two different illuminations.
Then, we trained the network in a siamese configuration, for which we again imposed both outputs to be similar to the reference one.
In addition, we introduced the following loss function:
\begin{equation}
\label{eq_ssimloss}
\mathcal{L}_{SSIM}(\textbf{y}_1,\textbf{y}_2,\textbf{y}^*) = SSIM(\textbf{y}_1,\textbf{y}_2)
\end{equation}
which is the structural similarity (SSIM) \cite{wang2004image}, usually employed to measure how similar two images are.
\textcolor{black}{
This second part of the training was performed, during 10 epochs of the training data (40k triplets), 
requiring about 6 hours of training with the same parameters as in previous Section.
}

\subsection{Training the LSTM network}
After we obtain a good CNN, the second part of the training is designed to increase the stability of the outputs, given that we are processing sequences of consecutive images. The goal is to provide not only more meaningful images, but also fulfill the brightness constancy assumption.
For that purpose, we trained the whole DNN, including the LSTM network, with sequences of two consecutive images (i.e., taken at consecutive poses on a trajectory) under slightly different illumination conditions, while the reference ones presented the same brightness.
The loss function consists of the LogRMSE loss function \eqref{eq_logloss} to ensure that both outputs are similar to their respective reference ones, and the SSIM loss \eqref{eq_ssimloss} without the structural term (as images do not belong to the exact same place) between the two consecutive outputs to ensure that they have a similar appearance.
\textcolor{black}{
The LSTM training was performed during 10 epochs of the data (40k triplets), in about 12 hours 
with the same parameters as in previous Section.
}

\newcommand{\figa}{workshop}
\newcommand{\figf}{bear614}
\newcommand{\figg}{bear914}
\newcommand{\figh}{bluefox67}
\newcommand{\figi}{desk}
\newcommand{\figk}{sancho2}
\newcommand{\fign}{flicker2}
\begin{figure*}[!htb]	
	\centering
	\captionsetup[subfigure]{labelformat=empty,position=top}
	\subfloat[Original Input]
	{\includegraphics[width=\wresults]{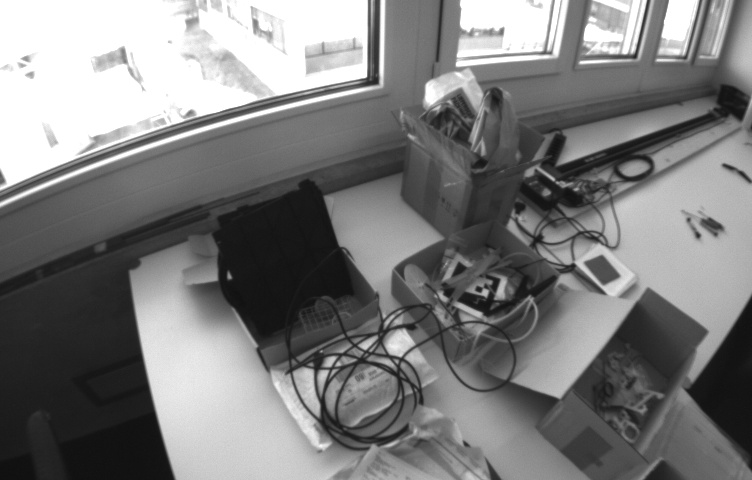}}	
	\quad
	\subfloat[FT-CNN Output]
	{\includegraphics[width=\wresults]{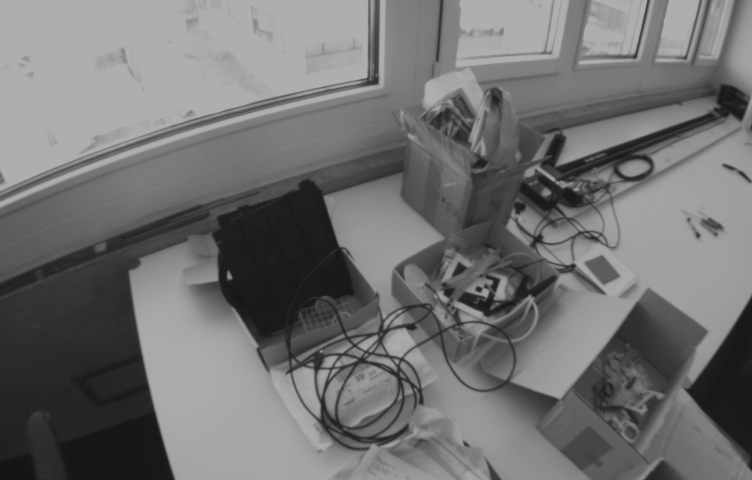}}		
	\quad
	\subfloat[FT-CNN Grad. Diff.]
	{\includegraphics[width=\wresults]{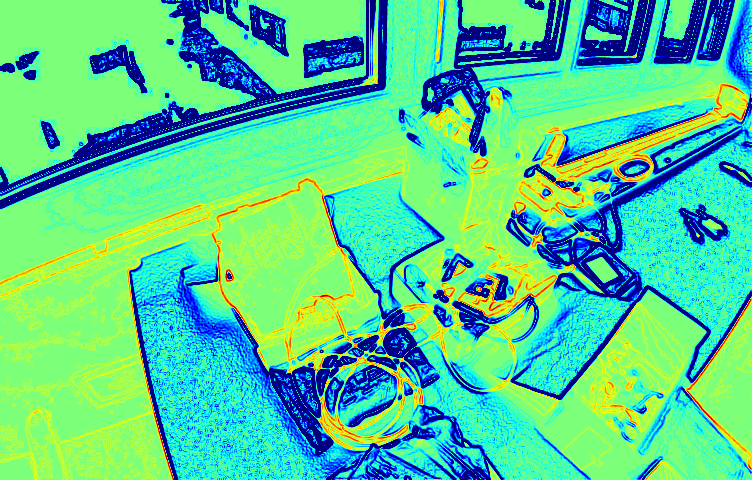}}		
	\quad
	\subfloat[FT-LSTM Output]
	{\includegraphics[width=\wresults]{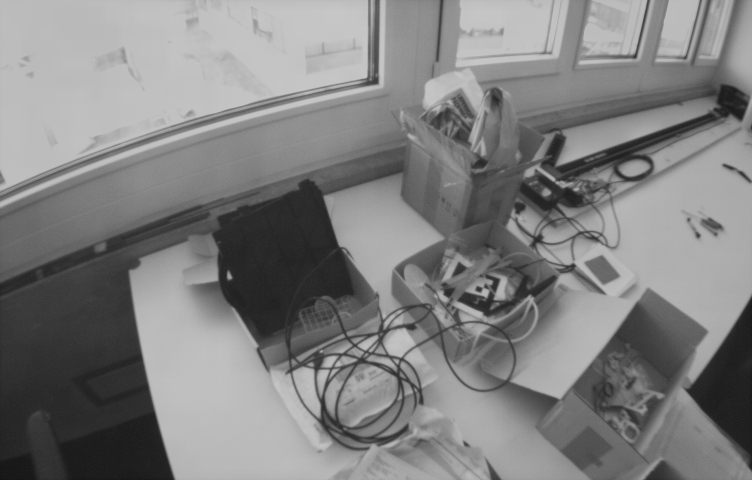}}	
	\quad
	\subfloat[FT-LSTM Grad. Diff.]
	{\includegraphics[width=\wresults]{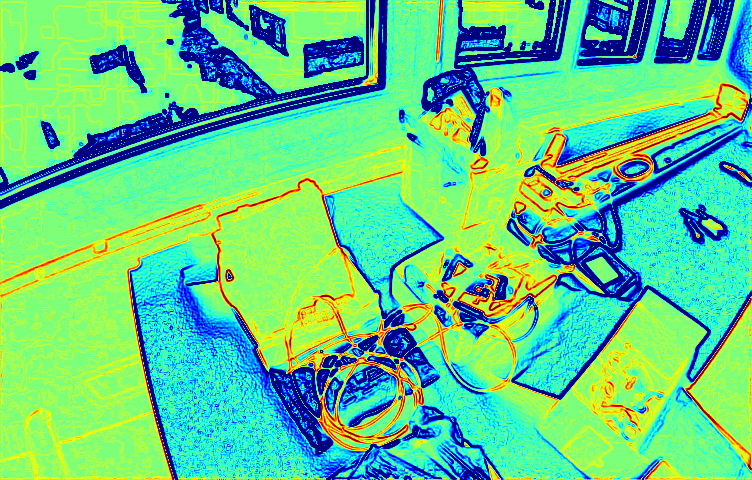}}		
	\quad
	\subfloat[Small-CNN Output]
	{\includegraphics[width=\wresults]{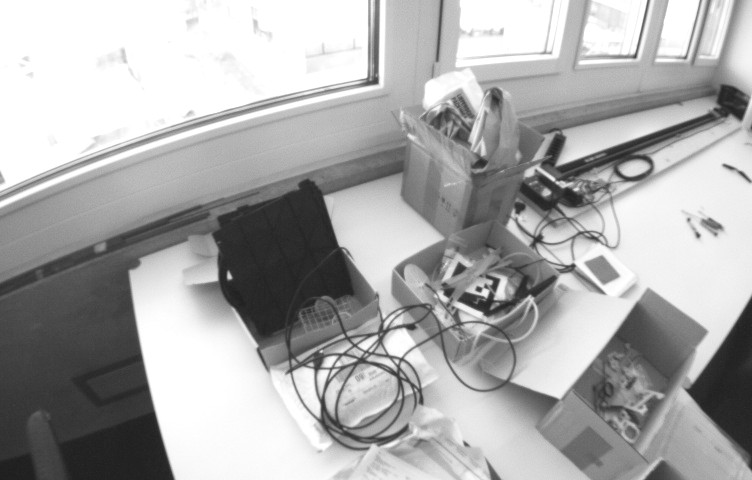}}	
	\quad
	\subfloat[Small-CNN Grad. Diff.]
	{\includegraphics[width=\wresults]{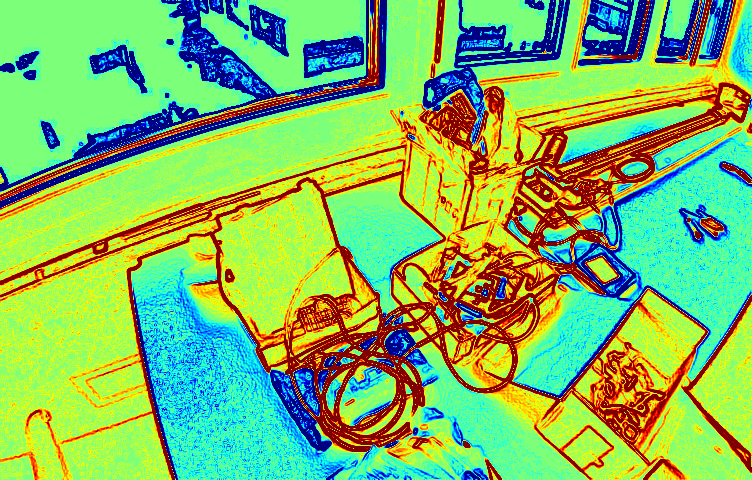}}	
	\\
	\vspace{\indfigd}
	\subfloat[]
	{\includegraphics[width=\wresults]{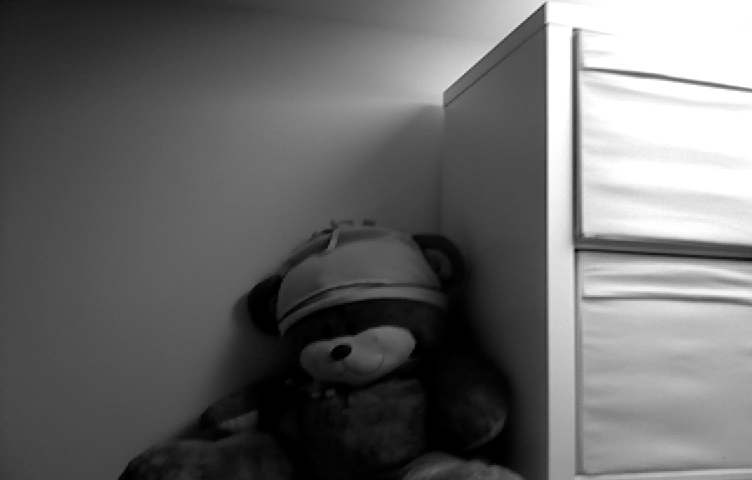}}	
	\quad
	\subfloat[]
	{\includegraphics[width=\wresults]{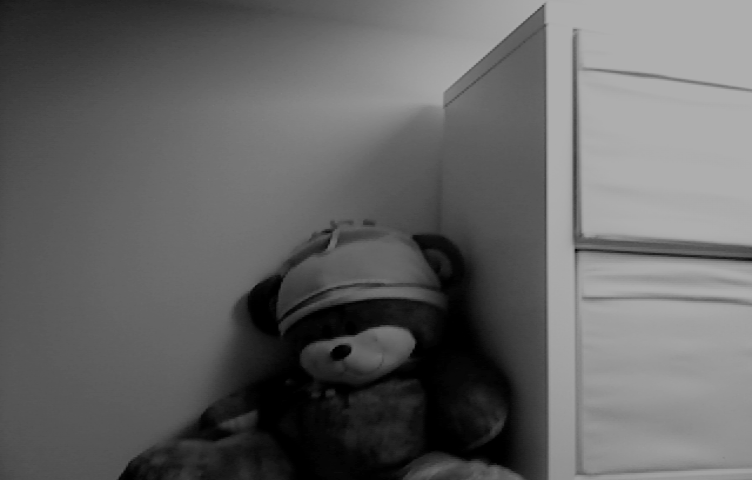}}		
	\quad
	\subfloat[]
	{\includegraphics[width=\wresults]{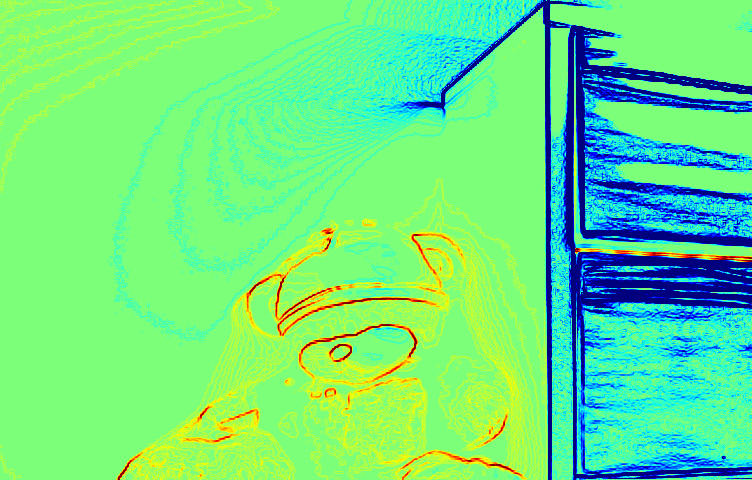}}		
	\quad
	\subfloat[]
	{\includegraphics[width=\wresults]{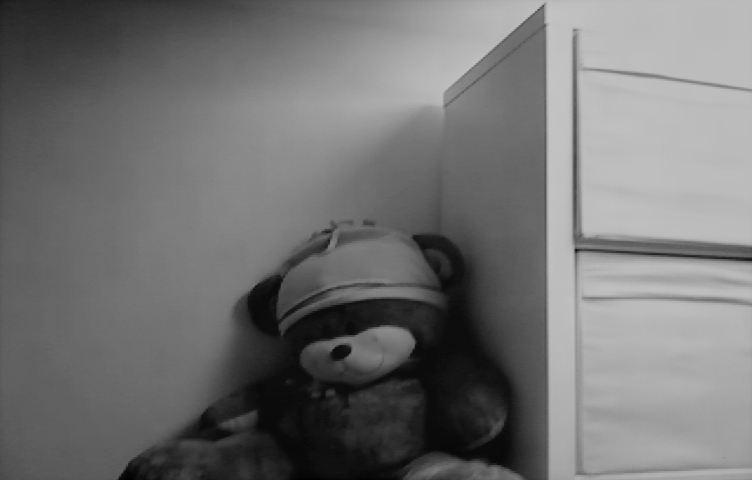}}	
	\quad
	\subfloat[]
	{\includegraphics[width=\wresults]{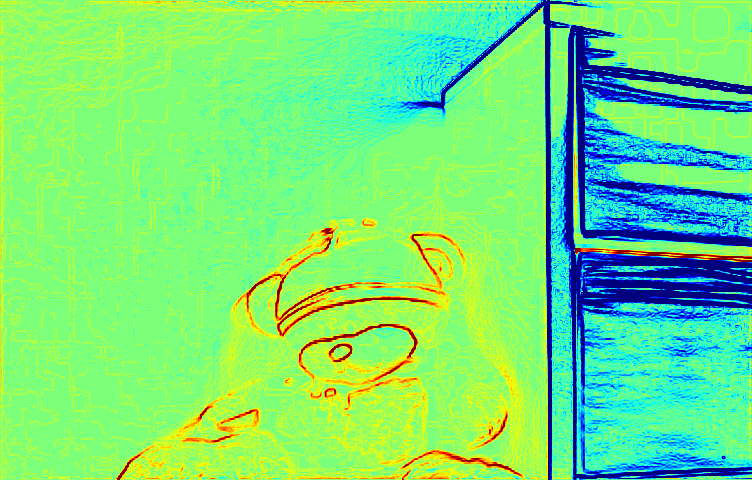}}		
	\quad
	\subfloat[]
	{\includegraphics[width=\wresults]{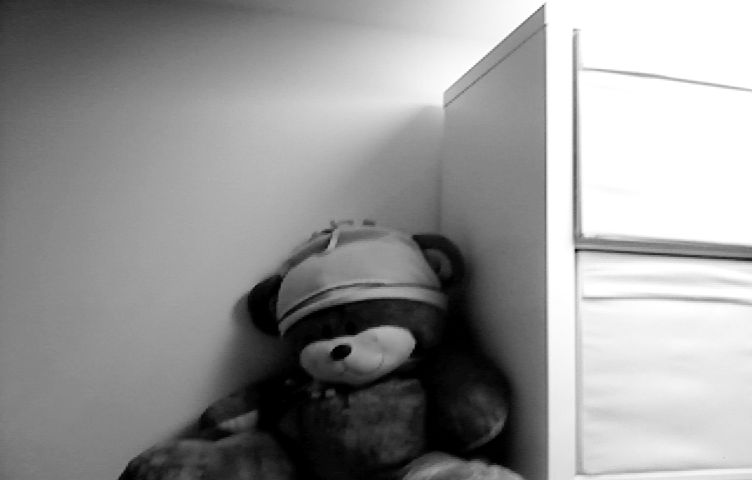}}	
	\quad
	\subfloat[]
	{\includegraphics[width=\wresults]{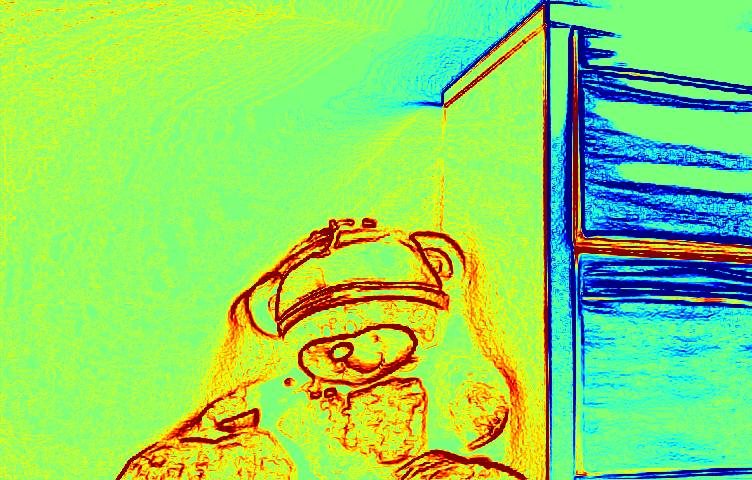}}	
	\\
	\vspace{\indfigd}	
	\subfloat[]
	{\includegraphics[width=\wresults]{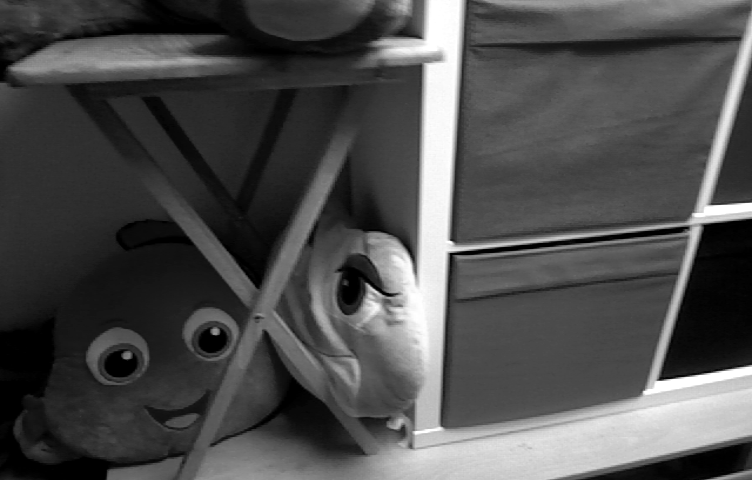}}	
	\quad
	\subfloat[]
	{\includegraphics[width=\wresults]{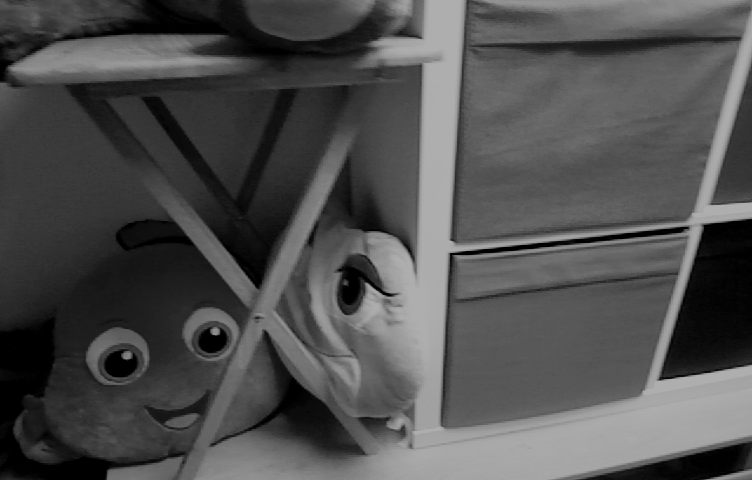}}		
	\quad
	\subfloat[]
	{\includegraphics[width=\wresults]{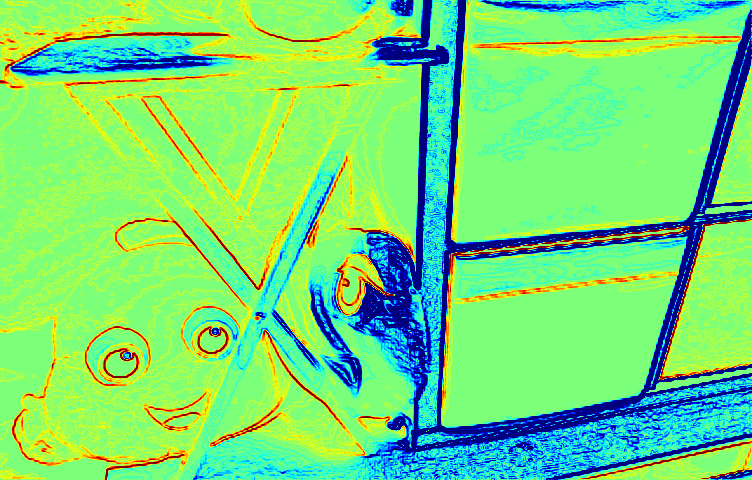}}		
	\quad
	\subfloat[]
	{\includegraphics[width=\wresults]{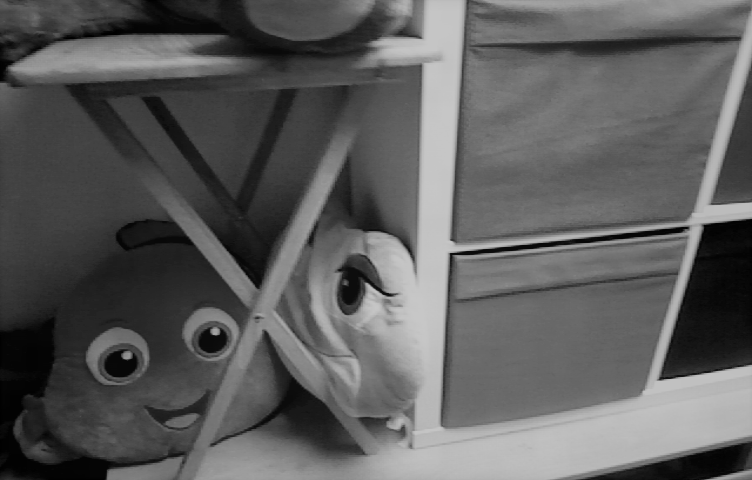}}	
	\quad
	\subfloat[]
	{\includegraphics[width=\wresults]{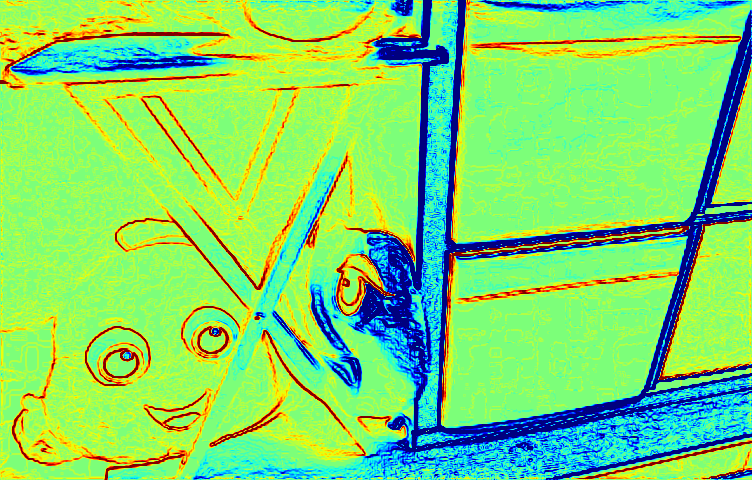}}		
	\quad
	\subfloat[]
	{\includegraphics[width=\wresults]{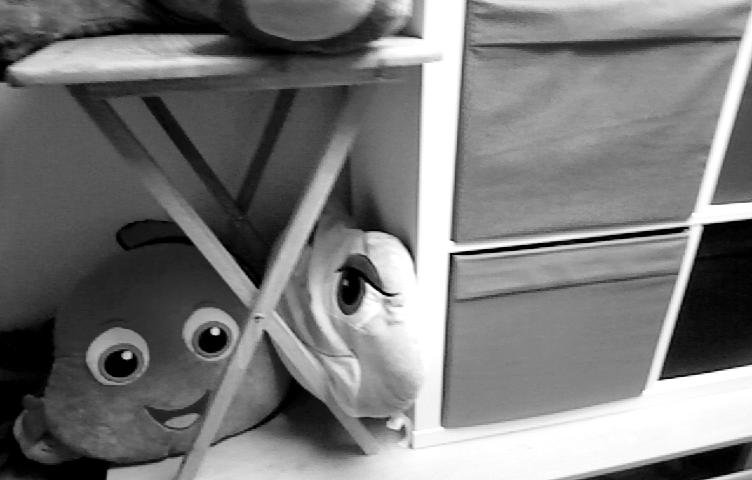}}	
	\quad
	\subfloat[]
	{\includegraphics[width=\wresults]{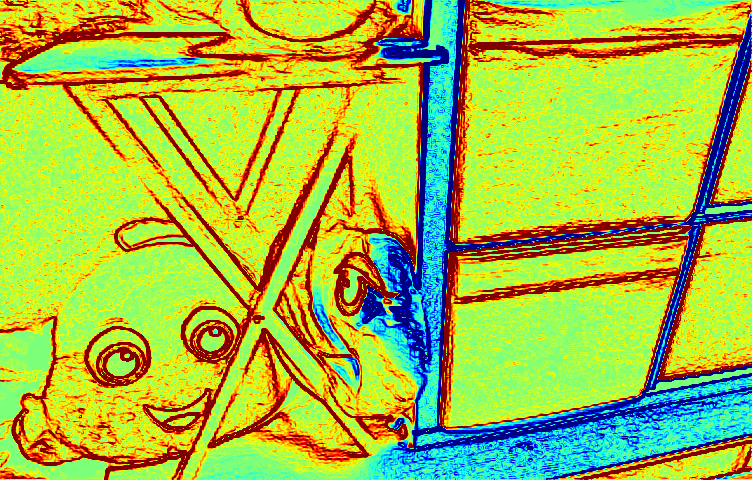}}	
	\\
	\vspace{\indfigd}
	\subfloat[]
	{\includegraphics[width=\wresults]{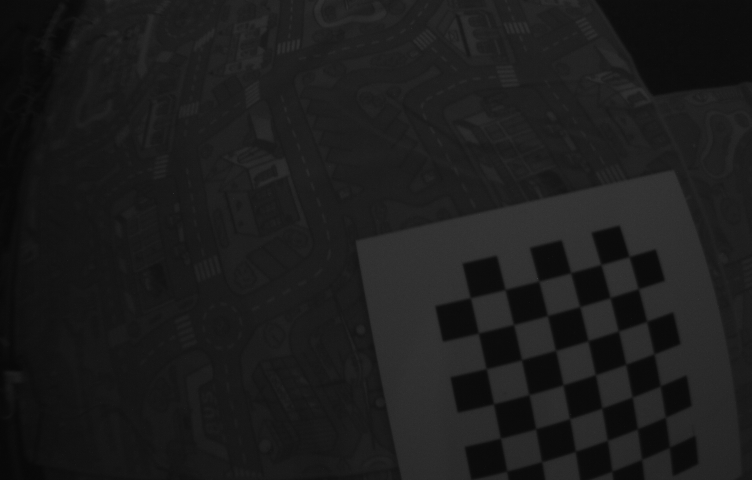}}	
	\quad
	\subfloat[]
	{\includegraphics[width=\wresults]{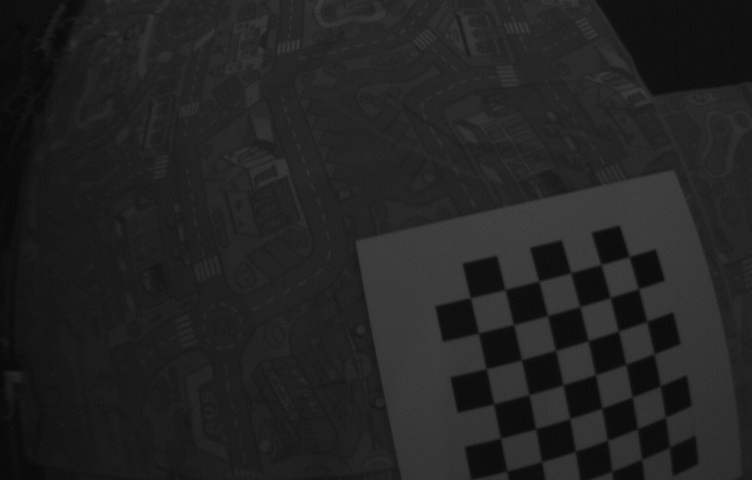}}		
	\quad
	\subfloat[]
	{\includegraphics[width=\wresults]{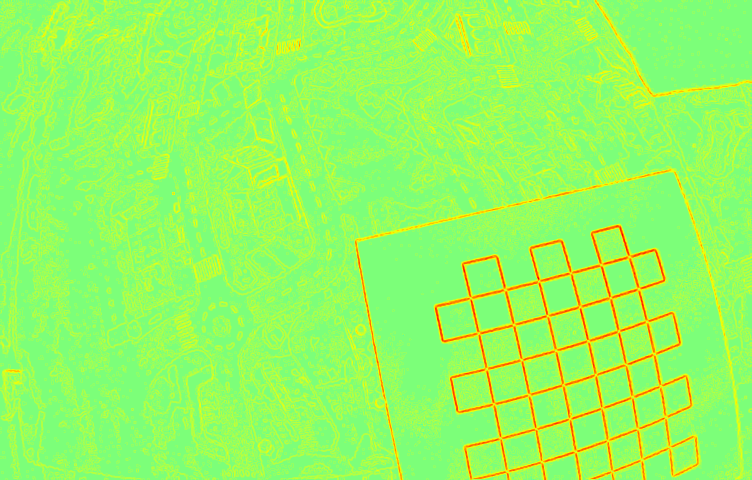}}		
	\quad
	\subfloat[]
	{\includegraphics[width=\wresults]{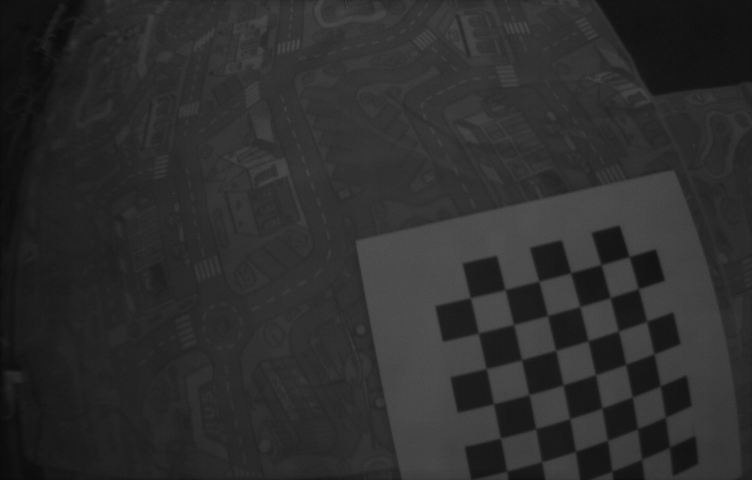}}	
	\quad
	\subfloat[]
	{\includegraphics[width=\wresults]{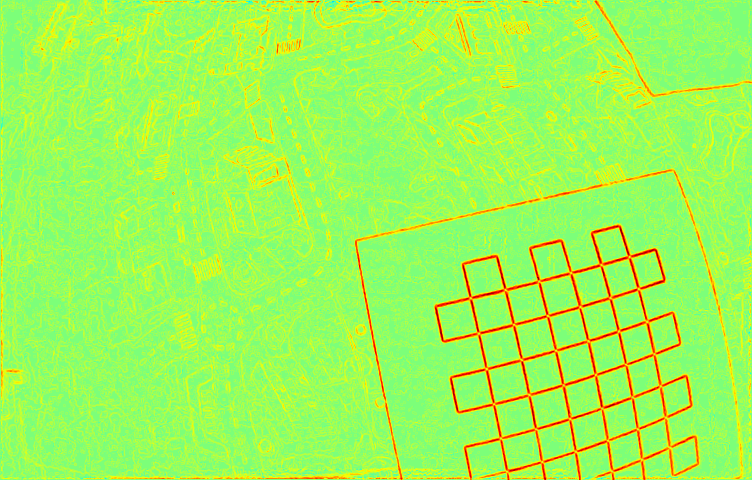}}		
	\quad
	\subfloat[]
	{\includegraphics[width=\wresults]{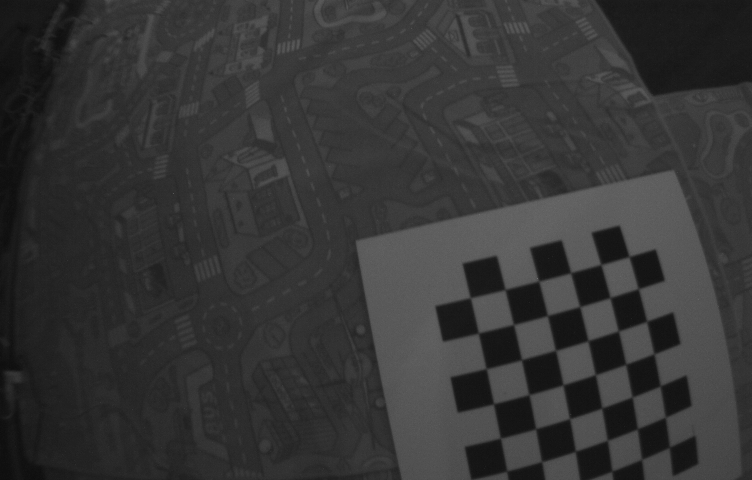}}	
	\quad
	\subfloat[]
	{\includegraphics[width=\wresults]{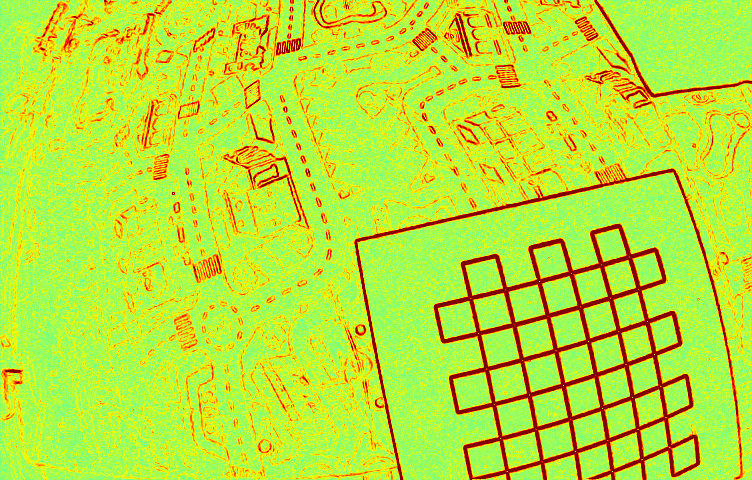}}	
	\\
	\vspace{\indfigd}
	\subfloat[]
	{\includegraphics[width=\wresults]{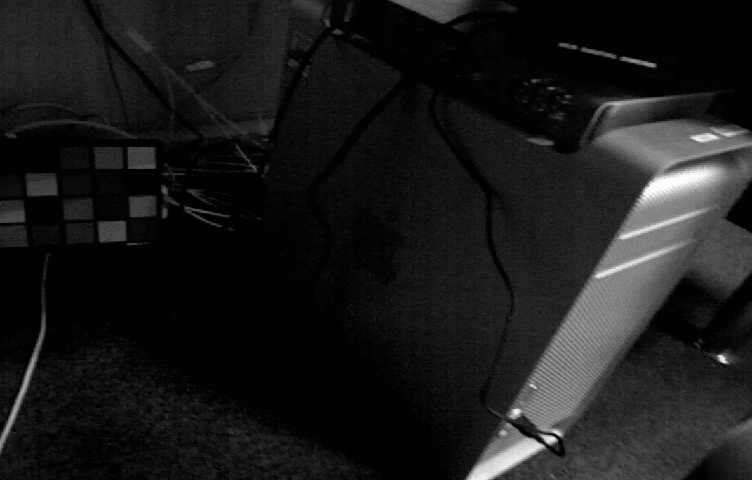}}	
	\quad
	\subfloat[]
	{\includegraphics[width=\wresults]{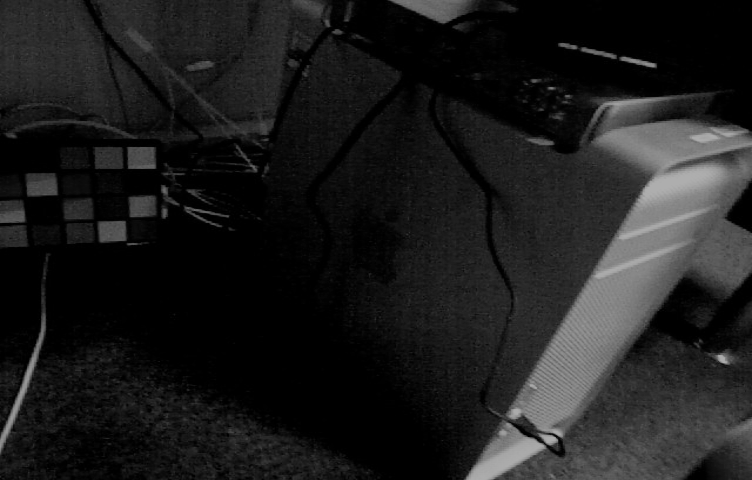}}		
	\quad
	\subfloat[]
	{\includegraphics[width=\wresults]{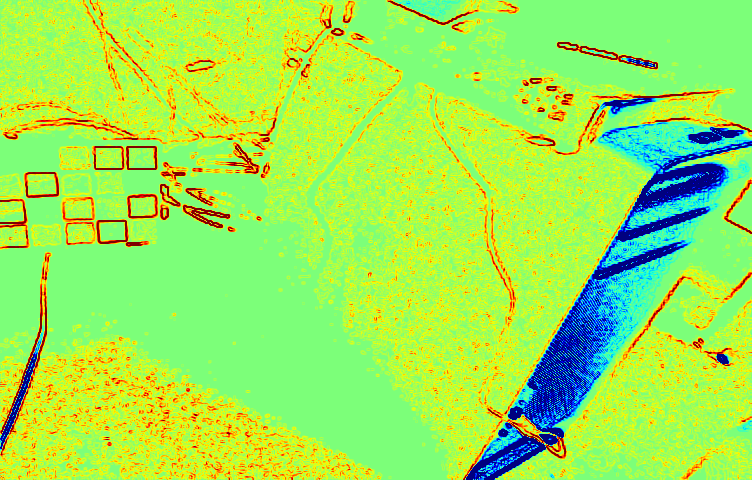}}		
	\quad
	\subfloat[]
	{\includegraphics[width=\wresults]{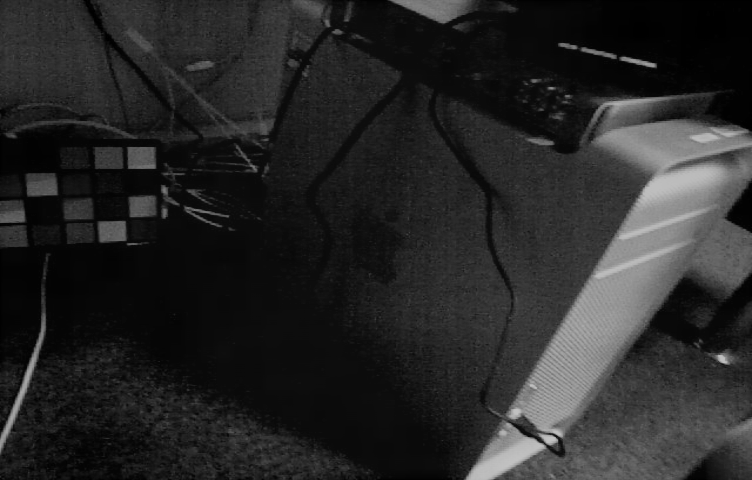}}	
	\quad
	\subfloat[]
	{\includegraphics[width=\wresults]{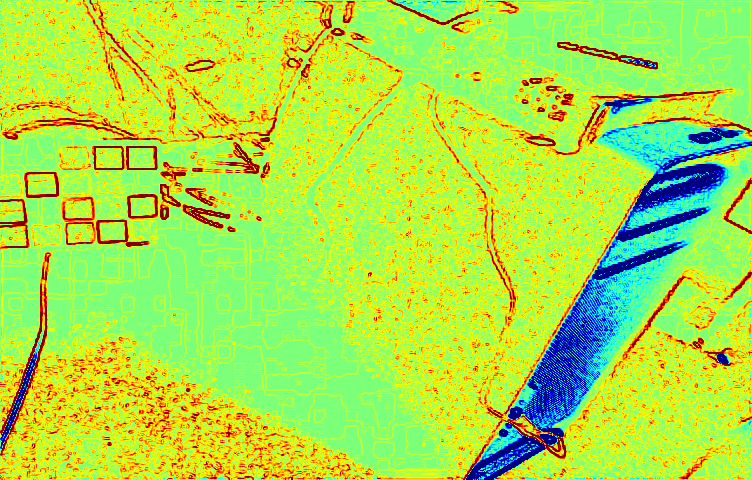}}		
	\quad
	\subfloat[]
	{\includegraphics[width=\wresults]{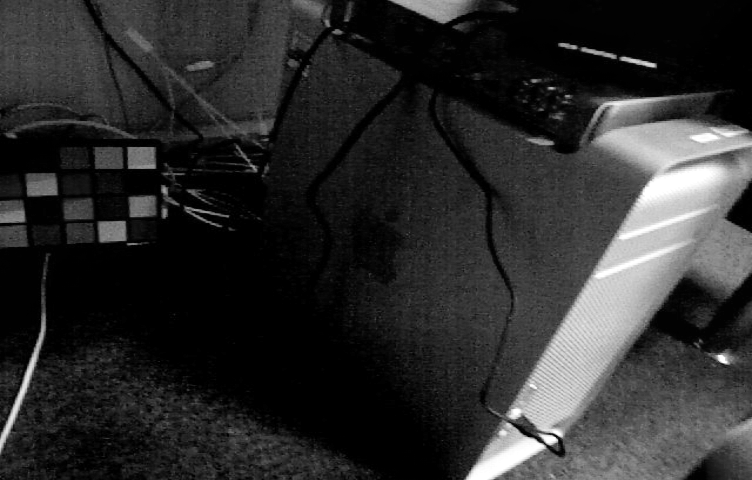}}	
	\quad
	\subfloat[]
	{\includegraphics[width=\wresults]{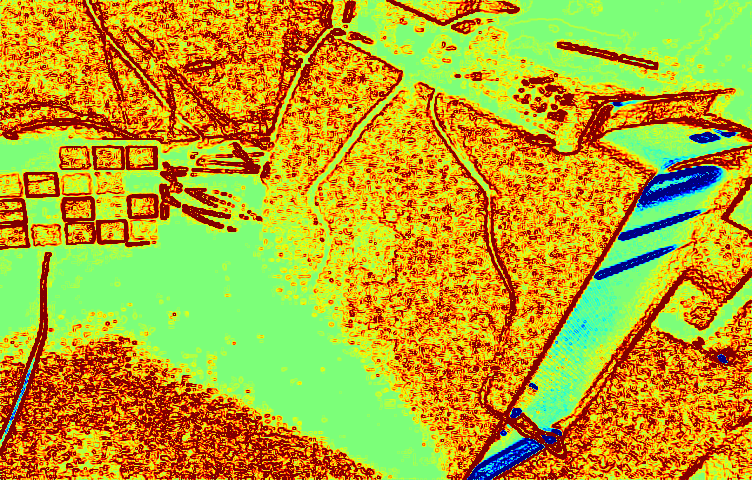}}	
	\\
	\vspace{\indfigd}	
	\subfloat[]
	{\includegraphics[width=\wresults]{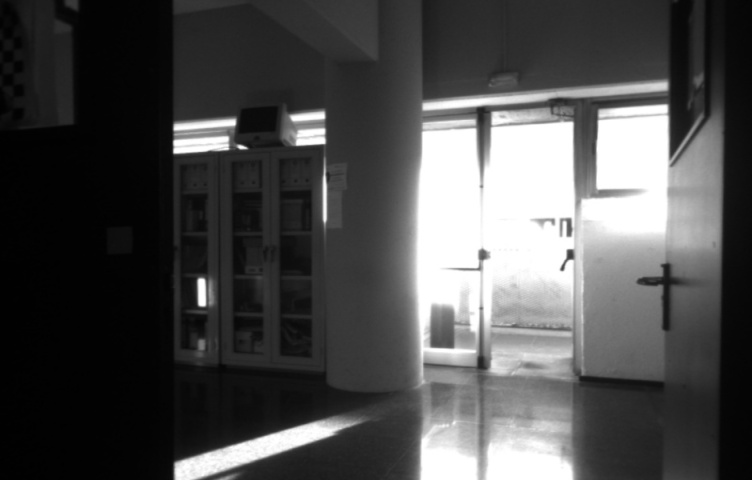}}	
	\quad
	\subfloat[]
	{\includegraphics[width=\wresults]{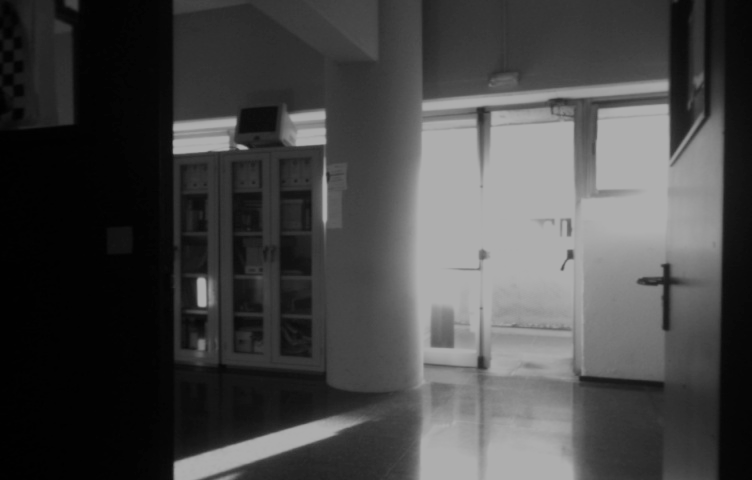}}		
	\quad
	\subfloat[]
	{\includegraphics[width=\wresults]{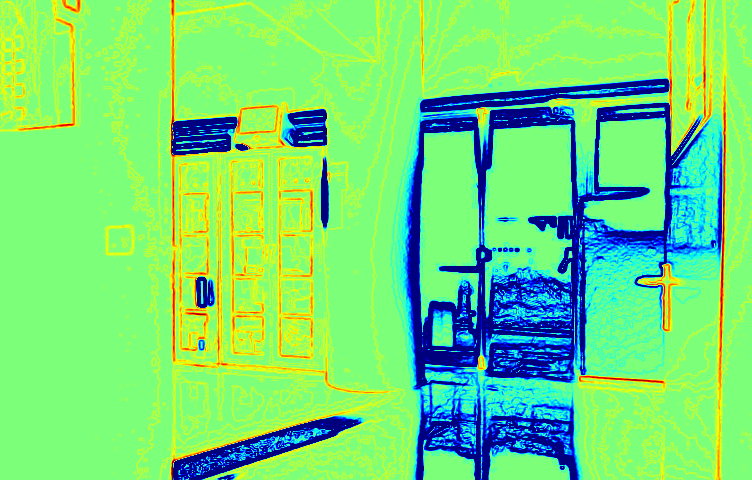}}		
	\quad
	\subfloat[]
	{\includegraphics[width=\wresults]{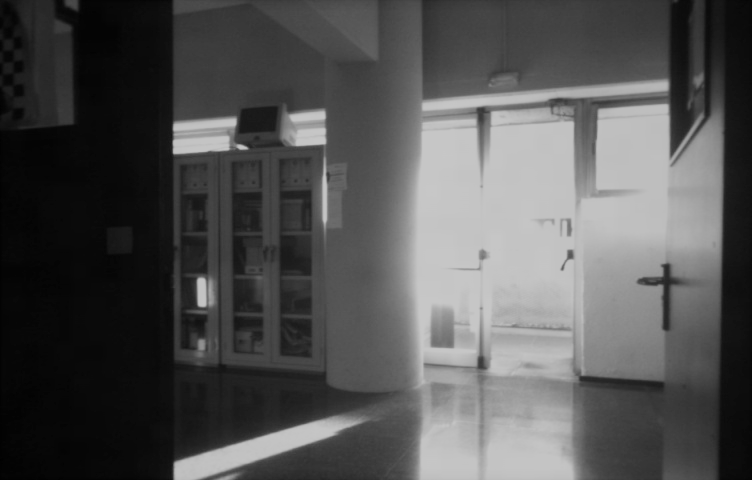}}	
	\quad
	\subfloat[]
	{\includegraphics[width=\wresults]{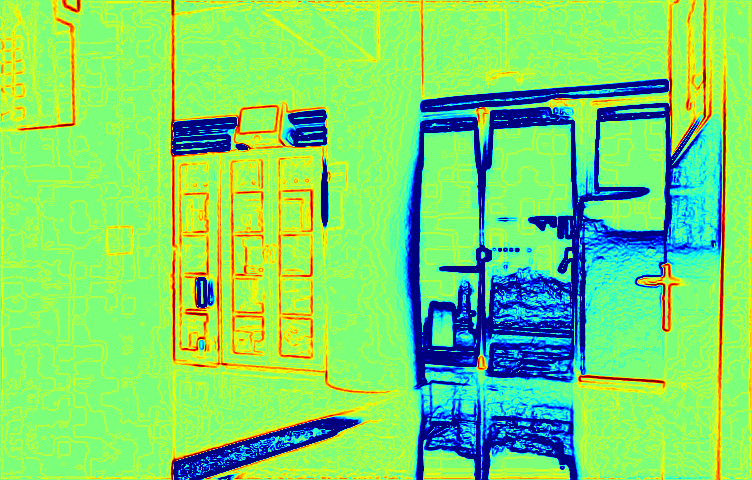}}		
	\quad
	\subfloat[]
	{\includegraphics[width=\wresults]{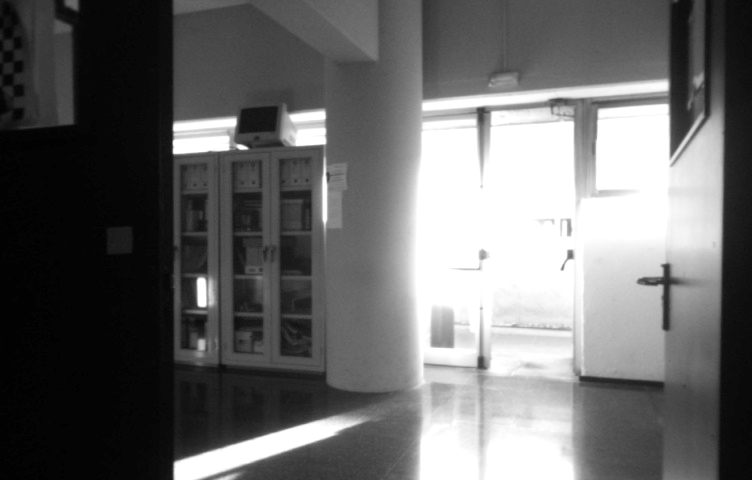}}	
	\quad
	\subfloat[]
	{\includegraphics[width=\wresults]{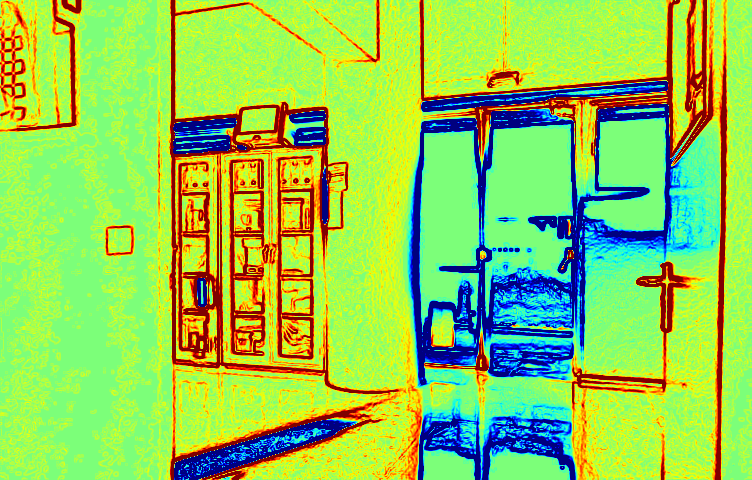}}	
	\\
	\vspace{\indfigd}
	\subfloat[]
	{\includegraphics[width=\wresults]{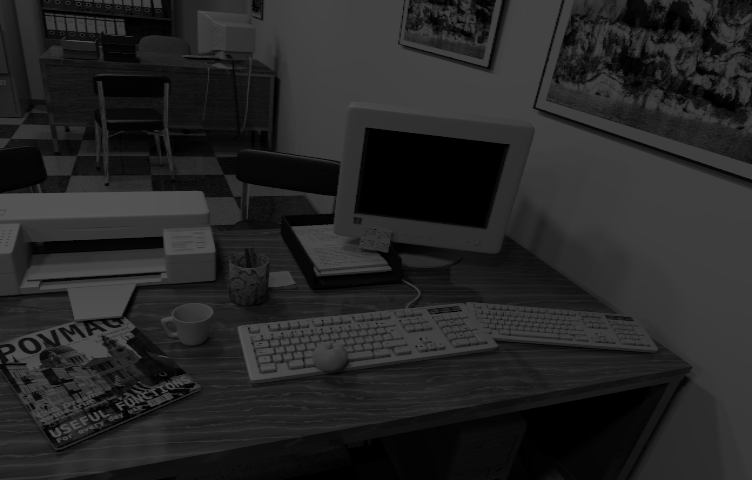}}	
	\quad
	\subfloat[]
	{\includegraphics[width=\wresults]{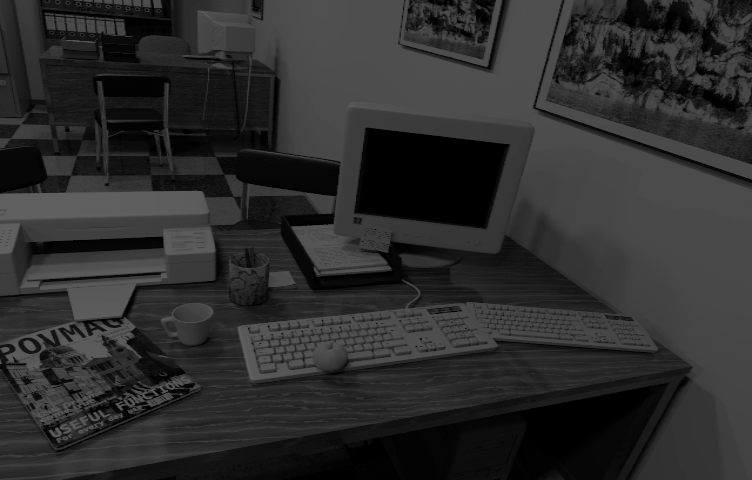}}		
	\quad
	\subfloat[]
	{\includegraphics[width=\wresults]{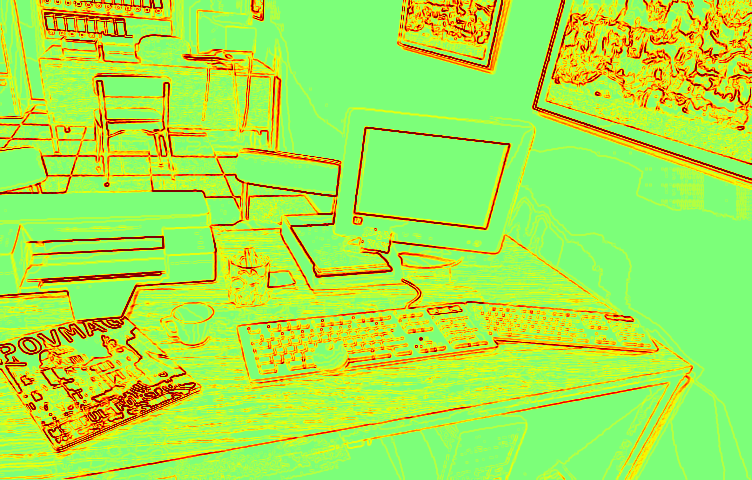}}		
	\quad
	\subfloat[]
	{\includegraphics[width=\wresults]{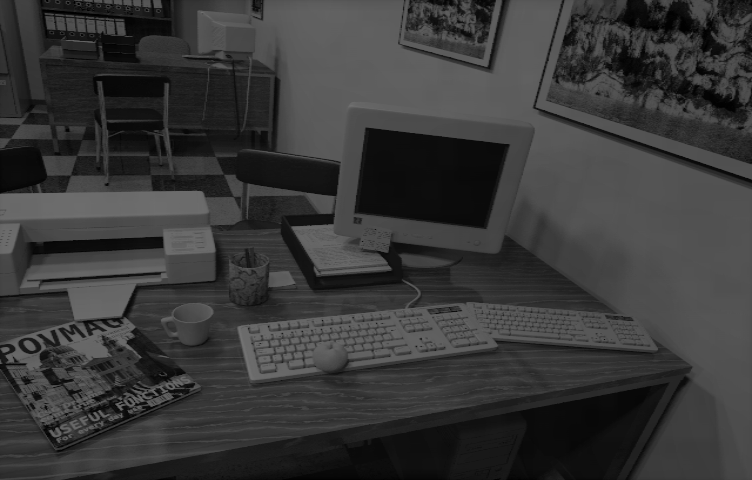}}	
	\quad
	\subfloat[]
	{\includegraphics[width=\wresults]{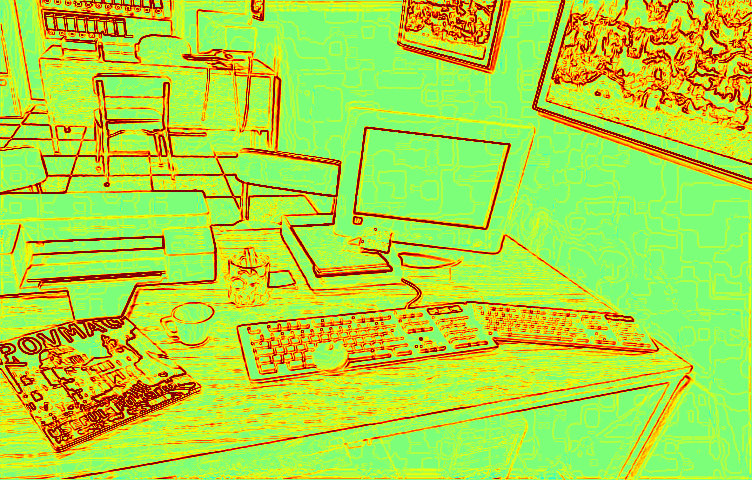}}		
	\quad
	\subfloat[]
	{\includegraphics[width=\wresults]{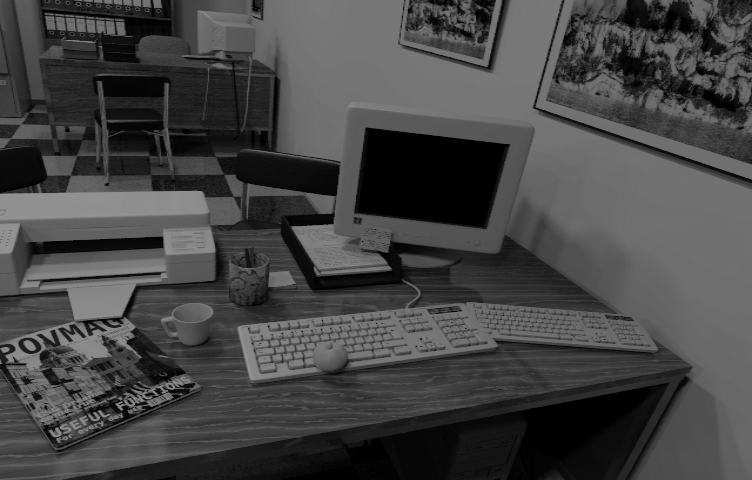}}	
	\quad
	\subfloat[]
	{\includegraphics[width=\wresults]{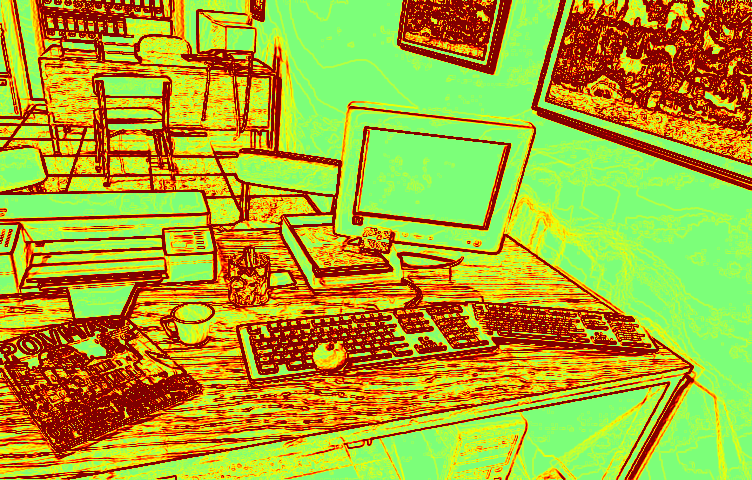}}	
	\\
	\caption{
		Outputs from the trained models and difference between the gradient images in some challenging samples extracted from the evaluation sequences (the scale for the \textit{jet} colormap remains fixed for each row).
	}
	\vspace{\indfigc}
	\label{fig_results}
\end{figure*}
\section{Experimental Validation}
\label{sec_experiments}
In this section, we evaluate the performance of our approach by measuring two different metrics: the increments of gradient magnitude in the processed images and the improvements in accuracy and performance of ORB-SLAM \cite{mur2015orb} and DSO \cite{engel2016direct}, two state-of-art VO algorithms for both \textit{feature-based} and \textit{direct} approaches, respectively.
For that, we first run the VO experiments with the original image sequence, several standard image processing approaches, 
i.e. Normalization (N),
Global Histogram Equalization (G-HE) \cite{russ1994image}, and
Adaptative Histogram Equalization (A-HE) \cite{zuiderveld1994contrast}.
\textcolor{black}{
Then, we also evaluate the VO algorithms with the image sequences produced with the trained networks: 
the fine-tuned approaches FT-CNN and FT-LSTM, and the reduced model trained from scratch Small-CNN.
}
%
Notice that, even though the CNN networks proposed in this paper (not the FT-LSTM) have been trained only with synthetic images with reduced size ($256\times180$ and $160\times120$ pixels for the fine-tuned and our proposal respectively), the experiments have been performed with full-resolution ($752\times480$ pixels) and real images.

\subsection{Gradient Inspection}

As stated before, one way of measuring the quality of an image is its amount of gradient.
Unfortunately, there is no standard metric for measuring the gradient information; actually, it is highly dependent on the application.
In the case of visual odometry, it is even more important, as most approaches are based on edge information (which is directly related to the gradient magnitude image).
\fig{fig_results} presents the estimated images and the difference between the gradients of the output and the input images for several images from the trained models in different datasets.
For the representation we have used the colormap \textit{jet}, i.e. from blue to red, with $\pm30$ units of range (negative values indicate a decrease of the gradient amount). 
In general, we observe a general tendency in all models to reduce the gradient amount in the most exposed parts of the camera as they are less informative due to the sensor saturation, while increasing the gradient in the rest of the image.

\subsection{Evaluation with state-of-art VO algorithms}
\begin{table*}[!htb]
	\small
	\centering
	\caption[]{
		\textcolor{black}{ORB-SLAM \cite{mur2015orb} average RMSE errors (\% first row) normalized by the length of the trajectory and percentage of the sequence without loosing the tracking (second row). A dash means that the VO experiment failed without initializing.}}
	\label{tab_orb}
	\begin{tabular}{lc|ccc|cc|c}
		Dataset & ORB-SLAM \cite{mur2015orb} & N & G-HE & A-HE & FT-CNN & FT-LSTM & Small-CNN \\
		\hline
		\hline
		\multirow{2}{*}{\textit{1-light}} 			& 3.91   & 4.07   & -      & -      & 3.52   & \textbf{3.49}   & 4.62   \\
													& 24.80  & 26.98  & -      & -      & 23.84  & 25.32  & \textbf{80.52}  \\
		\hline
		\multirow{2}{*}{\textit{2-lights}} 			& 2.19   & 2.17   & -      & 2.27   & \textbf{2.07}   & 2.09   & 2.72   \\
													& 68.92  & 68.76  & -      & 65.88  & 70.94  & \textbf{72.98}  & 68.76  \\
		\hline
		\multirow{2}{*}{\textit{3-lights}} 			& 3.78   & 3.81   & -      & 3.63   & \textbf{3.52}   & 3.81   & 3.65   \\
													& 100.00 & 100.00 & -      & 100.00 & 100.00 & 100.00 & 100.00 \\
		\hline
		\multirow{2}{*}{\textit{switch}} 			& 3.60   & 4.85   & -      & 4.56   & 5.64   & \textbf{2.66}   & 2.97   \\
													& 13.76  & 24.98  & -      & 8.84   & 7.32   & \textbf{31.02}  & 21.62  \\
		\hline
		\multirow{2}{*}{\textit{hdr1}} 				& 5.67   & 5.67   & 3.71   & -      & 5.22   & 5.21   & \textbf{4.77}   \\
													& 74.30  & 76.6   & 49.36  & -      & \textbf{81.54}  & 81.14  & 78.76  \\
		\hline
		\multirow{2}{*}{\textit{hdr2}} 				& 3.49   & 4.08   & 4.42   & 3.52   & \textbf{3.42}   & 3.88   & 3.51   \\
													& 74.86  & 70.50  & 34.12  & 25.3   & 74.52  & 71.02  & \textbf{75.22}  \\
		\hline
		\multirow{2}{*}{\textit{overexposed}}		& 2.64   & 2.57   & 2.59   & \textbf{2.53}   & 2.72   & 2.65   & 2.83   \\
													& 100.00 & 100.00 & 100.00 & 100.00 & 100.00 & 100.00 & 100.00 \\
		\hline
		\multirow{2}{*}{\textit{bright-switch}} 	& 3.13   & 3.08   & 2.03   & 3.10   & \textbf{1.97}   & 2.02   & 1.95   \\
													& 34.60  & 34.94  & 100.00 & 35.42  & 100.00 & 100.00 & 100.00 \\
		\hline
		\multirow{2}{*}{\textit{low-texture}} 		& -      & -      & -      & -      & \textbf{5.28}   & -      & -      \\
													& -      & -      & -      & -      & \textbf{39.08}  & -      & -      \\
	\end{tabular}
	\vspace{\indtaba}
\end{table*}
\begin{table*}[!htb]
	\footnotesize
	\centering
	\caption[]{
		\textcolor{black}{
			DSO \cite{engel2016direct} average RMSE errors normalized by the length of the trajectory 
			for each method and trained network when evaluating. A dash means that the VO experiment failed.}}
	\label{tab_dso}
	\begin{tabular}{lc|ccc|cc|c}
		Dataset & DSO \cite{engel2016direct} & N & G-HE & A-HE & FT-CNN & FT-LSTM & Small-CNN \\
		\hline
		\hline
		\textit{1-light} 	 	& 2.39 & -    & 2.37 & 2.42 & \textbf{2.36} & \textbf{2.36} & 2.40 \\
		\textit{2-lights} 	 	& 2.12 & -    & \textbf{2.05} & 2.12 & 2.12 & 2.15 & 2.14 \\
		\textit{3-lights} 	 	& \textbf{2.65} & -    & 2.66 & 2.66 & 2.66 & 2.69 & 2.69 \\
		\textit{switch} 	 	& -    & -    & -    & -    & 4.38 & 4.39 & \textbf{2.90} \\
		\textit{hdr1} 	     	& 2.46 & 4.80 & 2.34 & 2.52 & 2.42 & \textbf{2.17} & 2.44 \\
		\textit{hdr2} 	     	& 1.28 & -    & 1.59 & 3.17 & 1.23 & \textbf{1.22} & 2.57 \\
		\textit{overexposed} 	& 1.61 & 1.60 & 1.64 & 1.62 & \textbf{1.58} & \textbf{1.58} & 1.60 \\
		\textit{bright-switch} 	& 4.51 & -    & 1.49 & 1.47 & 1.93 & \textbf{1.73} & 4.43 \\
		\textit{low-texture} 	& 3.22 & 2.67 & 2.76 & 3.22 & 3.22 & \textbf{3.14} & 3.21 \\
	\end{tabular}
	\vspace{\indtabb}
\end{table*}
%
%
%
In order to evaluate the trained models in challenging conditions, we recorded 9 sequences with a hand-held camera in a room equipped with an OptiTrack system that allows us to also record the ground-truth trajectory of the camera and evaluate quantitatively the results.
Each sequence was recorded for several illumination conditions: first with $1-3$ lights available in the room, then without any light, and finally by switching the lights on and off during the sequence.
\ruben{
It is worth noticing that, despite the numerous public benchmarks available for VO, they are usually recorded in good and static illumination conditions, therefore our approach barely improves the trajectory estimation.
}

\tab{tab_orb} shows the results of ORB-SLAM in all the sequences mentioned above.
Firstly, we observe the benefits of our approach as our methods clearly outperform the original input and the standard image processing approaches in the difficult sequences (\emph{1-light} and \emph{switch}), while also maintaining a similar performance in the easy ones (\emph{2-lights} and \emph{3-lights}).
As for the different networks, we clearly observe the better performance of FT-LSTM in the difficult sequences, although the reduced approach Small-CNN reports a good performance in the scene with the switching lights.

\begin{table}[!htb]
	\vspace{-3mm}
	\centering
	\footnotesize
	\caption[]{Average runtime and memory usage for each network}
	\label{tab_times}
	\begin{tabular}{l|cr|rr}
		\multicolumn{1}{c|}{DNN} & Res. (pixels) & \multicolumn{1}{c|}{Memory} & \multicolumn{1}{c}{GPU} \\ 
		\hline
		\hline
		FT-CNN 	  	& $256\times180$ &  371 MiB &  23.80 ms 	\\ 
		FT-CNN 	  	& $756\times480$ & 1175 MiB & 149.72 ms 	\\ 
		FT-LSTM   	& $756\times480$ & 3897 MiB & 275.24 ms 	\\ 
		Small-CNN	& $160\times120$ &  135 MiB &  \textbf{4.77 ms}  \\ 
		Small-CNN	& $756\times480$ &  373 MiB &  \textbf{48.4 ms}  \\ 
	\end{tabular}
	\vspace{\indtabc}
\end{table}
The results obtained with DSO are represented in \tab{tab_dso}. Since all the methods were successfully tracked, we omit the tracking percentage.
%
In terms of accuracy, we again observe the good performance of the reduced approach, Small-CNN, with the direct approach. However, its accuracy is worse in the \textit{bright-switch} sequence but it still performs similar to the original sequence.
\subsection{Computational Cost}
Finally, we evaluate the computational performance of the two trained networks. 
For that, we compare the performance of the CNN and the LSTM, for both the training and the runtime image resolutions.
%
\ruben{
All the experiments were run on a Intel(R) Core(TM) i7-4770K CPU @ 3.50GHz and 8GB RAM, and an NVIDIA GeForce GTX Titan (12GB).
}
%
\tab{tab_times} shows the results of each model and all possible resolutions.
We first observe that while obtaining comparable results to the fine-tuned model, the small CNN can perform 
faster (a single frame processing takes 3 times less than with FT-CNN and up to 5 times less than FT-LSTM for the resolution $756\times480$ ), and therefore is the closest configuration to a direct application in a VO pipeline.
%
%
It is also worth noticing the important impact of the LSTM layers in the performance, because they not only require a high computational burden but also double the size of the encoder network (a consecutive image pair is needed).

\vspace{-2mm}
\section{{Conclusions}}
\label{sec_conclusions}
In this work, we tackled the problem of improving the robustness of VO systems under challenging conditions, such as difficult illuminations, HDR environments, or low-textured scenarios.
For that, we solved the problem from a deep learning perspective, for which we proposed two different architectures, a very deep model that is capable of producing temporally consistent sequences due to the inclusion of LSTM layers, 
\ruben{and a small and fast architecture more suitable for VO applications.}
%
We propose a multi-step training employing only reduced images from synthetic datasets, which are also augmented with a set basic transformations to simulate different illumination conditions and camera parameters, as there is no ground-truth available for our purposes.
We then compare the performance of two state-of-art algorithms in monocular VO, ORB-SLAM \cite{mur2015orb} and DSO \cite{engel2016direct}, when using the normal sequences and the ones produced by the DNNs, showing the benefits of our proposals in challenging environments.
%

%
\bibliographystyle{ieeetr}
\bibliography{./biblio/biblio_cnn_short}
\end{document}